\documentclass[sigconf]{acmart}

\usepackage{multirow}

\AtBeginDocument{%
  }

\copyrightyear{2023}
\acmYear{2023}
\setcopyright{acmlicensed}\acmConference[MM '23]{Proceedings of the 31st
ACM International Conference on Multimedia}{October 29-November 3,
2023}{Ottawa, ON, Canada}
\acmBooktitle{Proceedings of the 31st ACM International Conference on
Multimedia (MM '23), October 29-November 3, 2023, Ottawa, ON, Canada}
\acmPrice{15.00}
\acmDOI{10.1145/3581783.3613762}
\acmISBN{979-8-4007-0108-5/23/10}

\begin{document}

\title{MSECNet: Accurate and Robust Normal Estimation for 3D Point Clouds by Multi-Scale Edge Conditioning}


\author{Haoyi Xiu}
\affiliation{%
  \institution{National Institute of Advanced Industrial Science and Technology (AIST)}
  \city{Tokyo}
  \country{Japan}
  }
\email{hiroki-shuu@aist.go.jp}

\author{Xin Liu}
\authornote{Corresponding authors}
\affiliation{%
  \institution{National Institute of Advanced Industrial Science and Technology (AIST)}
  \city{Tokyo}
  \country{Japan}
  }
\email{xin.liu@aist.go.jp}


\author{Weimin Wang}
\authornotemark[1]
\authornote{Also is affiliated with Dalian University of Technology}
\affiliation{%
  \institution{National Institute of Advanced Industrial Science and Technology (AIST)}
  \city{Tokyo}
  \country{Japan}
  }
\email{weimin.wang@aist.go.jp}

\author{Kyoung-Sook Kim}
\affiliation{%
  \institution{National Institute of Advanced Industrial Science and Technology (AIST)}
  \city{Tokyo}
  \country{Japan}
  }
\email{ks.kim@aist.go.jp}

\author{Masashi Matsuoka}
\affiliation{%
  \institution{Tokyo Institute of Technology}
  \city{Tokyo}
  \country{Japan}
  }
\email{matsuoka.m.ab@m.titech.ac.jp}

\renewcommand{\shortauthors}{Haoyi Xiu, Xin Liu, Weimin Wang, Kyoung-Sook Kim, \& Masashi Matsuoka}
\newcommand{\etal}{\textit{et al.}}
\newcommand{\R}{\mathbb{R}}

\begin{abstract}
Estimating surface normals from 3D point clouds is critical for various applications, including surface reconstruction and rendering. While existing methods for normal estimation perform well in regions where normals change slowly, they tend to fail where normals vary rapidly. 
To address this issue, we propose a novel approach called MSECNet, which improves estimation in normal varying regions by treating normal variation modeling as an edge detection problem. MSECNet consists of a backbone network and a multi-scale edge conditioning (MSEC) stream. The MSEC stream achieves robust edge detection through multi-scale feature fusion and adaptive edge detection. The detected edges are then combined with the output of the backbone network using the edge conditioning module to produce edge-aware representations.
Extensive experiments show that MSECNet outperforms existing methods on both synthetic (PCPNet) and real-world (SceneNN) datasets while running significantly faster. We also conduct various analyses to investigate the contribution of each component in the MSEC stream. Finally, we demonstrate the effectiveness of our approach in surface reconstruction.

\end{abstract}





\maketitle

\section{Introduction}
3D point clouds can be used for many applications such as surface reconstruction~\cite{kazhdan2006poisson}, rendering~\cite{gouraud1971continuous}, and virtual reality~\cite{garrido2021point}. An essential step for many of them is to estimate normals at each point, which is to infer the local surface orientation from point samples. Therefore, the quality of normals estimated has a significant impact on the performance of those applications.

A classical approach for normal estimation is to fit a geometric primitive such as a plane~\cite{hoppe1992surface} or surface~\cite{cazals2005estimating} to the local neighborhood/patch of each point. 
However, a key challenge of these methods is that various parameters (e.g., the patch size) must be set manually without knowing the noise level and feature distribution, and their performance significantly depends on the choice of those parameters. 
Recently, data-driven approaches based on deep learning have been proposed to address this issue. They take a patch as input and estimate the normal of the patch center using deep neural networks (DNNs). There are two main types of approaches. The first one directly regresses the normal vector from features extracted by DNNs~\cite{guerrero2018pcpnet}. The other one uses DNNs to predict per-point weights for a weighted least square fitting and obtains the normal vector by solving it~\cite{lenssen2020deep}.

A recurrent issue of normal estimation is that the estimation in regions where normals change rapidly is not reliable. The problem is made even more challenging in the presence of data corruptions such as noise and density variations.
While some studies attempt to address this issue with hand-crafted estimators~\cite{zhou2022refine}, metric learning~\cite{wang2022deep}, or graph convolutions~\cite{li2022graphfit}, they often require manual preprocessing or still rely on surface fitting at the end, which may smooth out details and requires careful selection of the fitting order.

\begin{figure}[t]
    \centering
    \includegraphics[width=0.95\linewidth]{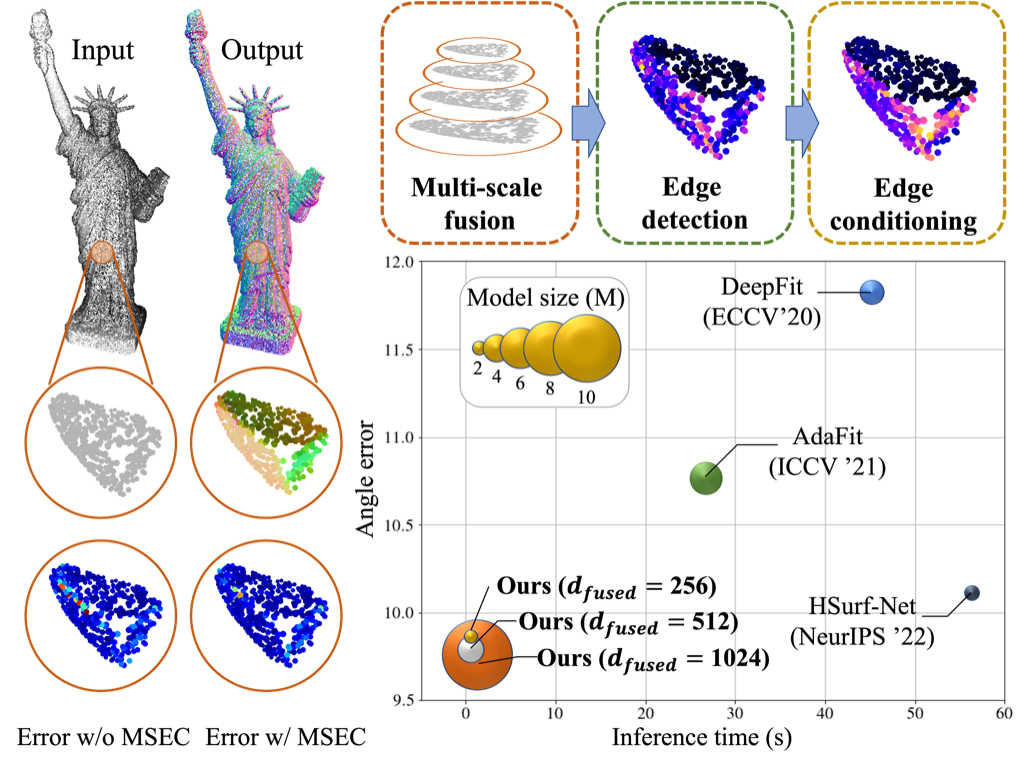}
    \caption{MSECNet performs accurate and robust normal estimation while running significantly faster than recent works. The MSEC stream improves estimation at sharp edges with multi-scale fusion, edge detection, and edge conditioning. }
    \label{fig: teaser}
\end{figure}

To address this issue, in this work, we propose Multi-Scale Edge Conditioning Network (MSECNet) for accurate and robust surface normal estimation\footnote{Code and pretrained models is available in \url{https://github.com/martianxiu/MSECNet}.}. 
The overview of our approach is presented in Fig.~\ref{fig: teaser}. 
MSECNet achieves accurate estimation at normal varying regions by the explicit modeling of normal variations. 
We treat the modeling of normal variations as the modeling of geometric edges. An edge, in essence, describes a variation in certain properties. 
A smooth change represents a weak edge while a drastic change (e.g., a corner) indicates a strong edge. 
Therefore, explicit modeling of geometric edges can lead to better modeling of normal variations, which in turn leads to more accurate estimation. 

To model geometric edges, we propose the MSEC stream which consists of multi-scale fusion, edge detection, and edge conditioning (see Fig.~\ref{fig: overview}).
The fusion step performs multi-scale feature fusion by reusing the multi-scale features of a backbone network. The fusion is necessary for successful edge detection as edges are multi-scale structures and the detection needs to be robust against various data corruption. 
Unlike previous works (e.g., ~\cite{zhu2021adafit} and \cite{li2022hsurf}) whose fusion aggregates information from larger scales by pooling, we distribute information from larger scales to the smallest one by interpolation. This is because we aim to detect edges that are spatially localized and aggregation destroys such information. 
The fused features are then used for adaptive edge detection in which the adaptability allows for the enhancement of true edges.
Lastly, the edge conditioning step merges the detected edges with the representation learned by the backbone to produce edge-aware features.

Extensive experiments demonstrate that MSECNet achieves state-of-the-art performance on the synthetic PCPNet dataset~\cite{guerrero2018pcpnet}. To investigate its generalizability, we apply pretrained MSECNet to the real-world SceneNN dataset~\cite{hua2016scenenn}. The results indicate that MSECNet outperforms all baselines by a large margin. 
Furthermore, we verify that MSECNet runs $\sim$61x faster than the previous works during inference. Moreover, we perform various analyses to show how each step of the MSEC stream helps. Finally, we demonstrate the utility of MSECNet in surface reconstruction.  

Our main contributions are summarized as follows:
\begin{itemize}
    \item We propose MSECNet, a new normal estimation approach that models normal variations by explicit edge modeling. 
    It achieves state-of-the-art performance on both synthetic and real-world datasets while showing strong robustness against different noise levels and density variations. Also, it runs significantly faster than cutting-edge networks.      
    \item We propose the MSEC stream that improves normal estimation at normal varying regions by the combination of multi-scale fusion, adaptive edge detection, and edge conditioning. We verify that the MSEC stream significantly advances the performance of the backbone network. 
\end{itemize}

\section{Related Work}

\subsection{Traditional Normal Estimation}
The classical methods for normal estimation are based on Principle Component Analysis (PCA)~\cite{hoppe1992surface,dey2006normal,alliez2007voronoi}. They perform a local covariance analysis on a local patch and use the direction of minimal variance as the normal. 
On the other hand, methods for surface reconstruction such as moving least squares~\cite{lancaster1981surfaces,alexa2001point}, osculating jets~\cite{cazals2005estimating},
and spherical fitting~\cite{guennebaud2007algebraic}
are also used for estimating normals. 
While their behavior is theoretically well-understood, they require careful selection of parameters such as the neighborhood size~\cite{mitra2003estimating} and often need special treatment when data are corrupted.

\subsection{Normal Estimation with DNNs}
One of the earliest approaches first projects a Hough accumulator into an image and then the normal is estimated by a convolutional neural network (CNN)~\cite{boulch2016deep}. Similarly, PointProNets~\cite{roveri2018pointpronets} and Refine-Net~\cite{zhou2022refine} convert a local point patch to a 2D height map that can be fed into CNN. 
However, they tend to lose fine geometric details due to the conversion.
To address this issue, some methods~\cite{guerrero2018pcpnet,sharma2021point} leverage PointNet~\cite{qi2017pointnet}, which directly takes point clouds as input, and aggregates points into a single vector from which the normal is regressed.
This type of approach is further improved in accuracy and efficiency by introducing multi-scale structures~\cite{guerrero2018pcpnet}, a point-voxel architecture~\cite{hashimoto2019normal}, and local plane constraint~\cite{zhou2020normal}. 

On the other hand, some studies leverage intrinsic problem structures to improve data efficiency. For instance, Lenssen et al.~\cite{lenssen2020deep} use a graph neural network to produce per-point weights and then use them for a weighted least-square plane fitting. Similarly, DeepFit~\cite{ben2020deepfit} adopts PointNet to predict weights for jet fitting. 
Both of them try to improve the robustness of estimation and data efficiency with the fitting techniques.
However, the fitting tends to smooth the surfaces. Besides, the polynomial order of the jet fitting is difficult to determine because the complexity of the surfaces varies greatly.
AdaFit~\cite{zhu2021adafit} addresses this issue by predicting an additional offset to adjust point distributions.
Zhang et al.~\cite{zhang2022geometry} learns the optimal set of points from which the normal vector is regressed.
Zhou~\etal~\cite{zhou2023improvement} simplifies the surface fitting by focusing on critical points found by top-K selection.
HSurf-Net~\cite{li2022hsurf} proposes a hypersurface fitting that frees the user from choosing the polynomial order. NeAF~\cite{li2023NeAF}, on the other hand, learns the angle field by predicting the offset between the random input vector and the ground truth normal.          
However, they tend to be slow in inference because they estimate only a single vector from a patch.

A recurrent issue of existing approaches is that they produce unreliable estimations at sharp edges. Hence, a line of research focuses on addressing this problem. Some approaches aim to accurately predict normal vectors in areas where the normal varies by separately modeling sharp and non-sharp patches~\cite{lu2020deep,wang2022deep}, but these methods require manual preprocessing. Other approaches use graph convolutions to learn more about local geometry~\cite{lenssen2020deep,li2022graphfit}, but these techniques have limited access to multiple scales and may be less robust to data corruption. Additionally, many of these methods still rely on surface fitting, which can smooth out details. MESCNet differs from these methods in that it does not require any preprocessing or surface fitting. Furthermore, it can directly access features at all scales, allowing for robust and accurate estimation.



\subsection{Edge Detection}
Computational edge detection is an extremely well-studied topic. Pioneering works such as Sobel detector~\cite{kittler1983accuracy}, zero-crossing~\cite{marr1980theory}, and Canny method~\cite{canny1986computational} detect edges using gradients. However, these methods usually cannot cope with the complexity of real-world applications. To address this issue, learning-based edge detection methods (e.g., \cite{martin2004learning, arbelaez2010contour, dollar2014fast}) are proposed. Nevertheless, their performance depends on manual processing and expert knowledge.     
Recently, edge detectors based on DNNs have shown impressive performance. For instance, HED~\cite{xie2015holistically} performs edge detection using a cascaded architecture in which one layer produces a scale-specific edge prediction. 
Edge maps of different scales are then combined to produce accurate edge predictions.
Building on top of HED, RCF~\cite{liu2017richer} makes full use of multi-scale features by using all convolutional layers for prediction. BDCN~\cite{he2019bi} further enhances the prediction with scale-specific supervision. 
PiDiNet~\cite{su2021pixel} integrates traditional operators to convolutions for efficient detection. RINDNet~\cite{pu2021rindnet} models different types of edges with dedicated decoders.
EDTER~\cite{pu2022edter} exploits the full image context with transformers. 

A key observation is that multi-scale features are essential for successful edge detection because edges are multi-scale structures~\cite{marr1980theory}. Regarding multi-scale fusion strategies for normal estimation, some studies use multiple patches with different patch sizes as input and build separate networks for each scale~\cite{guerrero2018pcpnet,ben2019nesti}. However, creating separate networks for each scale is costly. Another approach is to aggregate large-scale information through max-pooling and then copy the resulting vector for all points~\cite{zhu2021adafit,li2022graphfit,li2022hsurf}. Our approach differs from theirs in that we distribute information instead of aggregating it, since we aim to detect spatially localized edges. Also, their fusion is limited to adjacent scales, while ours has direct access to all scales, leading to a more accurate and robust representation.


\section{Multi-Scale Edge Conditioning}\label{sec: method}

\begin{figure*}[tbhp]
    \centering
    \includegraphics[width=0.8\linewidth]{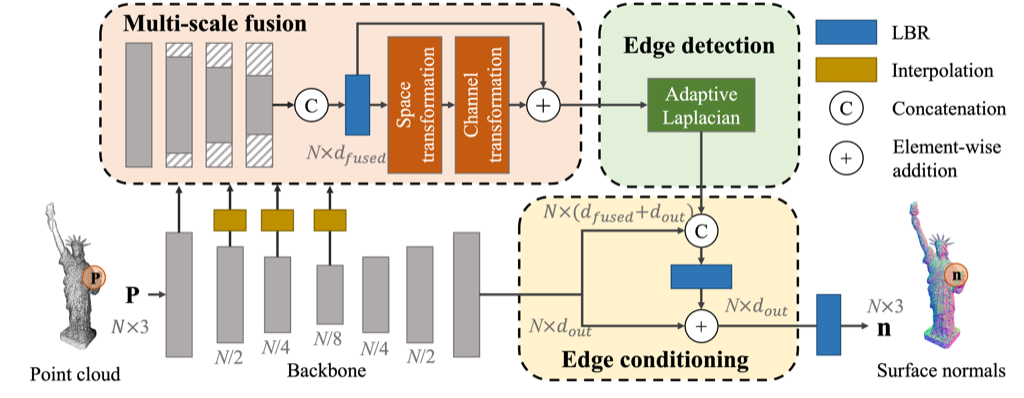}
    \caption{The architecture of MSECNet. The MSEC stream consists of multi-scale fusion, edge detection, and edge conditioning steps. LBR stands for a sequential application of a Linear transformation, a batch normalization, and a ReLU. }
    \label{fig: overview}
    \Description{}
\end{figure*}

Let $\textbf{P} \in \R^{N\times3}$ be a patch consisting of $N$ points and $\textbf{p}_i=(x_i, y_i, z_i)\in\R^{3}$ denote the coordinates of point $i$. A point $i$ can be associated with a feature $\textbf{f}_i \in \R^{d}$. Our task is to estimate the unoriented normals $\textbf{n}\in\R^{N \times 3}$ of this patch from $\textbf{P}$. 
The overall architecture is illustrated in Fig.~\ref{fig: overview}. 
Given a patch of point clouds, a backbone extracts multi-scale features for each point. 
Then, features from each stage are fed into the MSEC stream. The motivation for introducing the MSEC stream is explained in Sec.~\ref{sec: motivation}.
The first step of the stream is feature fusion (Sec.~\ref{sec: multi-scale fusion}) which merges features from different scales. The resulting multi-scale features are subsequently used for adaptive edge detection (Sec.~\ref{sec: adaptive edge detection}). Then, the detected edges are used to condition the output of the backbone to produce edge-aware representation (Sec.~\ref{sec: edge conditioning}).  
Finally, normals are regressed from the resulting features. Other details are elaborated in Sec.~\ref{sec: arch}.

Unlike previous work that estimates only one normal for a patch, we estimate normals for all points of the patch. This allows for a significant speedup in inference, which is described in Sec.~\ref{sec: arch}. 
            
\subsection{Motivation}~\label{sec: motivation}
Successful edge modeling must rely on multi-scale analysis for two reasons. First, edges are naturally multi-scale~\cite{marr1980theory}, and thus combining multiple scales improves edge detection~\cite{ren2008multi}. Second, single-scale edge detection is hardly effective in the presence of data corruption such as noise and density variations. Combining information from large and small scales can help to estimate normals accurately while being robust against data corruption.

To obtain features of different scales, it is natural to use hierarchical DNNs (e.g., PointNet++~\cite{qi2017pointnet++}), which perform efficient multi-scale learning through the use of local aggregation and subsampling. One might think that the output of such networks could be used for edge detection. However, from an edge detection perspective, the problem with these networks is that each layer only has direct access to adjacent scales. Direct access to non-adjacent scales is hampered by intermediate processing such as transformations and subsampling, which can severely compromise the rich information contained in the original features. Therefore, we construct a new stream so that the edge detector can directly access all scales.

\subsection{Multi-Scale Fusion}\label{sec: multi-scale fusion}
Effectively combining features of different scales is critical for edge detection. In this work, a feature from one stage of the backbone network represents a single-scale feature that characterizes a particular scale based on receptive field size and density. 

A common multi-scale fusion strategy for normal estimation aggregates information from larger scales using max-pooling to extract more information about the underlying geometry~\cite{zhu2021adafit,li2022graphfit,li2022hsurf}. 
However, since edges are spatial structures and aggregation destroys spatial relationships, such fusion methods are not suitable for our purpose. 
Therefore, instead of aggregating from larger scales, we distribute information from larger scales to preserve spatial structures.  
As shown in Fig.~\ref{fig: overview}, an output from a stage undergoes a 3-neighbor interpolation before entering the MSEC stream. 
The interpolation ensures that the spatial density of the feature is the same as the input, since we want to detect edges from the raw point cloud. Unlike the popular gradual interpolation strategy (e.g., PointNet++~\cite{qi2017pointnet++}), our strategy allows a jump across the hierarchy. This jump results in ``blurred'' point clouds that efficiently describe multi-scale structures. As a result, each point has direct access to rich features learned at different scales.


Let $\textbf{f}^s_i$ denote the feature of point $i$ whose scale is $s \in \{1,2,...,S\}$. Given $S$ features from different scales, the initial multi-scale feature is produced by merging all features of different scales and embedding them into a single vector: 
\begin{equation}\label{eq: initial fusion}
    \textbf{f}_i^{fused} = \phi \left(\left[\textbf{f}_i^1, \textbf{f}_i^2, ..., \textbf{f}_i^S\right]\right),
\end{equation}
where $\phi:\R^{d_1+ d_2 + ... + d_S} \to \R^{d_{fused}}$ denotes an LBR and $[\cdot]$ denotes the concatenation. $\phi$ is shared among points and nonlinearly maps features of different scales to the fixed $d_{fused}$-dimensional space. Hence, the resulting features have the accuracy of small scales and robustness of large scales.
However, a point-wise fusion is not sufficient for edge detection due to various factors: 1) small-scale features are too sensitive to noise; 2) large-scale features have poor localization; and 3) interpolation may produce biased multi-scale features due to biases in sampling density. 
A key observation is that the above issues can be addressed by spatial filtering.  
Therefore, we introduce space transformation that spatially filters the fused feature. Let $\mathcal{N}(i)$ denote the local neighborhood of point $i$. The space transformation is defined as: 
\begin{equation}\label{eq: space trans}
    \textbf{f}^{sp}_i = \max_{j\in\mathcal{N}(i)} \alpha \left(\left[\textbf{p}_j-\textbf{p}_i, \textbf{f}_j^{fused}\right]\right),
\end{equation}
where $\alpha:\R^{3+d_{fused}} \to \R^{d_{fused}}$ is an MLP consisting of point-wise linear transformations, batch normalizations, and ReLUs. $\max$ denotes a max-pooling operation within the neighborhood of point $i$. The relative positions are concatenated to $\textbf{f}^{fused}_i$ to better describe the 3D geometry of points. Consequently, the space transformation strives to address the above issues by sharpening important details on the one hand and smoothing harmful signals on the other. 
Furthermore, we introduce the channel transformation that performs intra-channel filtering:
\begin{equation}\label{eq: channel trans}
    \textbf{f}_i^{spch} = \beta \left(\textbf{f}^{sp}_i \right),
\end{equation}
where $\textbf{f}_i^{spch}$ indicates features that undergo spatial and channel transformations and $\beta:\R^{d_{fused}} \to \R^{d_{fused}}$ represents an MLP, similar to $\alpha$. 
With more refined and noise-free features as input, the channel transformation aims to produce more discriminative features by nonlinear mapping, in the same sense as a position-wise feed-forward network in the transformer~\cite{vaswani2017attention}.

Successive transformations inevitably increase the difficulty of optimization. Transformations may also result in the loss of useful information contained in the original features. 
Therefore, we adopt a residual structure~\cite{he2016deep} to ease the optimization while reducing information loss. 
Specifically, the output of the fusion is: 
\begin{equation}\label{eq: multi-scale fusion}
    \textbf{f}_i^{ms} = \textbf{f}^{fused}_i + \textbf{f}_i^{spch},
\end{equation}
where $\textbf{f}_i^{ms}$ denotes the resulting multi-scale features. 

\subsection{Adaptive Edge Detection}\label{sec: adaptive edge detection}


Given a multi-scale feature, the detection step transforms the features so that the output describes the ``edgeness'' of the point. For our purposes, the desired detector should satisfy two requirements. First, the operator should be computationally efficient, since it must be computed for many training iterations. Second, the detector should be permutation invariant because the point clouds can come in any order. 
To this end, we advocate the use of the discrete Laplacian (or the umbrella operator~\cite{taubin1995signal}).
Unlike sophisticated edge detectors (e.g., ~\cite{dollar2014fast}), the Laplacian is a linear operator and can be computed by simply taking the differences between neighbors. In addition, the Laplacian is permutation invariant because it involves only pairwise differences and a symmetric function (i.e., averaging).


However, there are situations where the Laplacian may not be sufficient or even problematic. For instance, Laplacian is known to be sensitive to noise and may amplify noisy signals, leading to less reliable edge detection. On the other hand, there might be weak edges that could be smoothed out easily, which is undesirable since these edges reflect small changes in surface normals that should be accurately modeled. To address these concerns, we propose introducing adaptability into the Laplacian to sharpen useful edges while suppressing noisy signals. Specifically, edge detection using the adaptive Laplacian can be expressed as follows:
\begin{equation}\label{eq: adaptive laplacian}
    \textbf{e}_i =\varphi \left(\frac{1}{|\mathcal{N}(i)|} \sum_{j\in\mathcal{N}(i)} \theta \left(\textbf{f}_j^{ms} - \textbf{f}_i^{ms}\right)\right), 
\end{equation}
where $\textbf{e}_i$ denotes the detected edges and $\theta,\varphi:\R^{d_{fused}} \to \R^{d_{fused}}$ denote point-wise MLPs. $\theta$ filters the pairwise difference in a local neighborhood to emphasize significant differences and smooth harmful ones. 
$\varphi$ is shared among all points to capture a more complex relationship between detected edges, making the representation more discriminative.   
Consequently, $\textbf{e}_i$ can effectively describe the ``edgeness'' of points. Note that Eq.~(\ref{eq: adaptive laplacian}) degenerates to the raw Laplacian when $\theta$ and $\varphi$ are removed.

\subsection{Edge Conditioning}\label{sec: edge conditioning}
While detected edges are accurate and robust to data corruption, the edges have limited information about surface orientation as the detector is based on the isotropic Laplacian.
Consider a scenario where two perpendicular planes intersect. Our edge detector would ideally produce zero vectors for every point within the planes. However, this would result in the detector being unable to distinguish between points on the different planes, which in turn would prevent the differentiation of their normal vectors.
To address this issue, we must supplement the detected edges with global information that contains surface orientation. This can be achieved by combining the detected edges and the output of the backbone. As the backbone network essentially learns the representation of the surface, it naturally contains information about surface orientations (not necessarily the globally correct one because our task is to estimate unoriented normals). 
Specifically, we propose an edge conditioning operation that conditions edges on the backbone output and generates an edge-aware surface representation:
\begin{equation}\label{eq: edge conditioning}
    \textbf{f}_i^{cond} = \textbf{f}^{b}_i + \gamma \left( \left[ \textbf{f}^{b}_i, \textbf{e}_i \right] \right),    
\end{equation}
where $\textbf{f}_i^{cond}$ denotes the edge-aware surface representation, $\textbf{f}_i^{b}\in\R^{d_{out}}$ is the backbone output, and $\gamma: \R^{d_{fused} + d_{out}} \to \R^{d_{out}}$ denotes an LBR. It conditions edges to global information with nonlinear mapping to construct a feature that can be used for accurate normal estimation. Similar to Eq.~(\ref{eq: multi-scale fusion}), we adopt a residual structure to facilitate optimization while reducing information loss.   

\subsection{MSECNet}\label{sec: arch}
\indent 

\textbf{Backbone.}
As shown in Fig.~\ref{fig: overview}, MSECNet consists of a backbone and an MSEC stream. We adopt the popular PointNet++~\cite{qi2017pointnet++} except that each encoder layer is adapted into a residual structure~\cite{he2016deep} to save computation. The network performs subsamplings three times in the encoder, and each time the number of points is halved while the feature dimension is doubled. Then, three upsampling layers are used to recover the original point cloud with nearest-neighbor interpolation. Every interpolation layer adopts a U-Net~\cite{unet}-like skip connection to help feature reconstruction. The feature dimension of the first stage is set to 64. 
Unlike many prior studies that generate a single normal for a patch center, we estimate the normals for all points in the patch.
In the case of a single-output network, it is necessary for the network to iterate through all points to compute normal vectors for the entire point cloud. 
In contrast, MSECNet only needs to run until the patches cover the whole cloud, making the inference dramatically faster (Sec.~\ref{sec: efficienncy}). 

\indent 

\textbf{Loss functions.}
We adopt two standard loss functions for normal estimation. The first one is the regression loss~\cite{guerrero2018pcpnet,zhang2022geometry}: 
\begin{equation}\label{eq: regression loss}
    \mathcal{L}^{reg} = \frac{1}{N}\sum_{i=1}^N \min(||\widehat{\textbf{n}}_i - \textbf{n}_i||^2, ||\widehat{\textbf{n}}_i + \textbf{n}_i||^2),
\end{equation}
where $\widehat{\textbf{n}}_i$ denotes an estimated normal. The regression loss is used to penalize the Euclidean distance between estimated and ground truth normals. Additionally, we adopt a sine loss~\cite{ben2020deepfit,li2022hsurf} to penalize the angular distance:
\begin{equation}\label{eq: sin loss}
    \mathcal{L}^{sin} = \frac{1}{N}\sum_{i=1}^N|\widehat{\textbf{n}}_i \times \textbf{n}_i|,
\end{equation}
where $\times$ denotes the cross product. Consequently, the total loss is
\begin{equation}\label{eq: total loss}
    \mathcal{L} = \mathcal{L}^{reg} + \mathcal{L}^{sin}. 
\end{equation}
We do not include other popular loss functions like the consistency loss~\cite{ben2020deepfit,zhu2021adafit,li2022graphfit} or geometry-guiding loss~\cite{zhang2022geometry,li2022hsurf} because they are not applicable or involves manual thresholding.

\begin{figure}[t]
    \centering
    \includegraphics[width=0.8\linewidth]{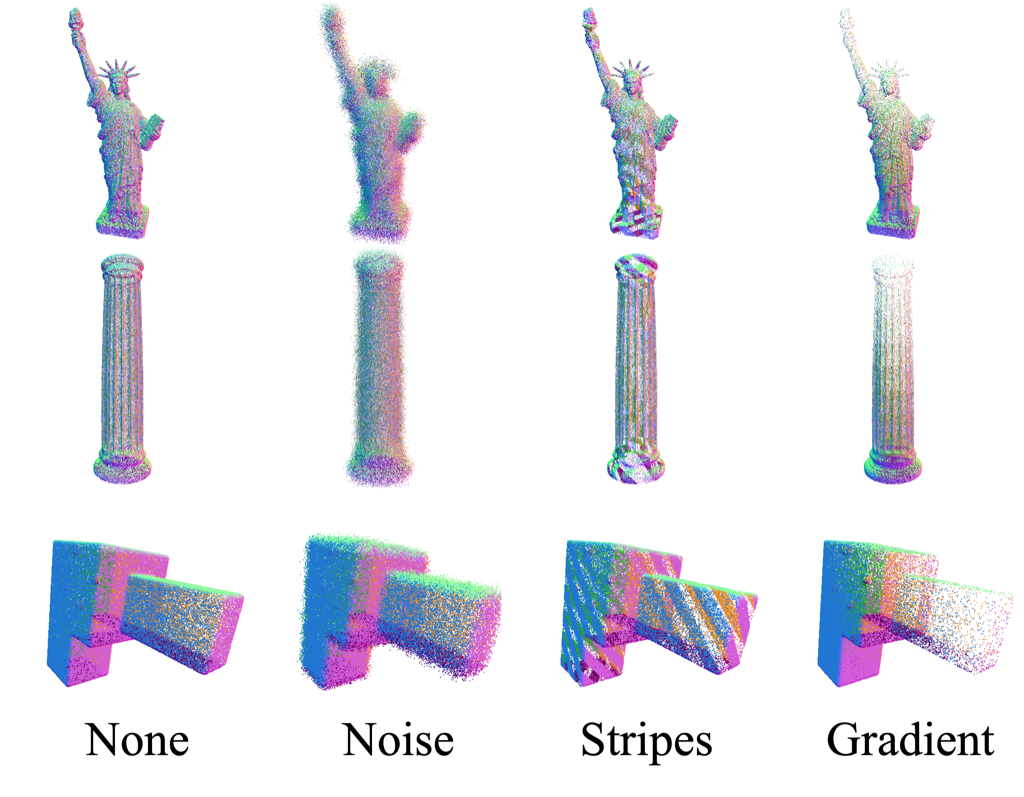}
    \caption{Examples of point clouds in the PCPNet dataset. 
    }
    \label{fig: pcpnnet example}
\end{figure}

\begin{table*}[t]
    \caption{Normal angle RMSE results on the PCPNet and SceneNN datasets. The lower the better.}
    \small
    \begin{tabular}{@{}l|c|lllllll|lll@{}}
    \toprule
        \multirow{3}{*}{Method} 
        & \multirow{3}{*}{Year} 
        & \multicolumn{7}{c|}{PCPNet Dataset}
        & \multicolumn{3}{c}{SceneNN Dataset}                              
        \\ 
        \cmidrule(l){3-12} 
        &
        & \multicolumn{4}{c|}{Noise $\sigma$}
        & \multicolumn{2}{c|}{Density}            
        & \multicolumn{1}{c|}{\multirow{2}{*}{Average}} 
        & \multicolumn{1}{c}{\multirow{2}{*}{Clean}} 
        & \multicolumn{1}{c|}{\multirow{2}{*}{Noise}} 
        & \multicolumn{1}{c}{\multirow{2}{*}{Average}} 
        \\ 
        \cmidrule(lr){3-8}
        &
        & None 
        & 0.12\% 
        & 0.6\% 
        & \multicolumn{1}{l|}{1.2\%} 
        & Stripes 
        & \multicolumn{1}{l|}{Gradient} 
        & \multicolumn{1}{c|}{}                         
        & \multicolumn{1}{c}{}                       
        & \multicolumn{1}{c|}{}                       
        & \multicolumn{1}{c}{}                         
        \\ 
        \midrule
        PCA~\cite{hoppe1992surface}
        & 1992
        & 12.29      
        & 12.87  
        & 18.38     
        & \multicolumn{1}{l|}{27.52}      
        &  13.66       
        & \multicolumn{1}{l|}{12.81}         
        & 16.25                     
        & 15.93                                     
        & \multicolumn{1}{l|}{16.32}            
        & 16.12    
        \\
        Jet~\cite{cazals2005estimating}
        & 2005
        & 12.35    
        & 12.84       
        & 18.33      
        & \multicolumn{1}{l|}{27.68}      
        & 13.39
        & \multicolumn{1}{l|}{13.13 }         
        & 16.29
        & 15.17
        & \multicolumn{1}{l|}{15.59}               
        & 15.38      
        \\
        PCPNet~\cite{guerrero2018pcpnet}
        & 2018
        & 9.64     
        & 11.51       
        & 18.27  
        & \multicolumn{1}{l|}{22.84}      
        & 11.73        
        & \multicolumn{1}{l|}{13.46}         
        &  14.58                                    
        & 20.86                                     
        & \multicolumn{1}{l|}{21.40}                
        & 21.13                                     
        \\
        Zhou~\etal~\cite{zhou2020normal} 
        & 2020
        & 8.67 
        &     10.49   
        &       17.62
        & \multicolumn{1}{l|}{24.14}      
        &         10.29
        & \multicolumn{1}{l|}{10.66}         
        &                          13.62                     
        &                               -             
        & \multicolumn{1}{l|}{-}                       
        &                      -                        
        \\
        Nesti-Net~\cite{ben2019nesti}
        & 2019
        & 7.06     
        &     10.24   
        &       17.77
        & \multicolumn{1}{l|}{22.31}      
        &         8.64
        & \multicolumn{1}{l|}{8.95}         
        &                         12.49                      
        &                              13.01              
        & \multicolumn{1}{l|}{15.19}                       
        &                          14.10                    
        \\
        Lenssen~\etal~\cite{lenssen2020deep}
        & 2020
        &     6.72 
        &        9.95
        &       17.18
        & \multicolumn{1}{l|}{21.96}      
        &         7.73
        & \multicolumn{1}{l|}{7.51}         
        &                         11.84                      
        &                              10.24              
        & \multicolumn{1}{l|}{13.00}                       
        &                          11.62                    
        \\
        DeepFit~\cite{ben2020deepfit}
        & 2020
        &     6.51 
        &        9.21
        &       16.73
        & \multicolumn{1}{l|}{23.12}      
        &         7.92
        & \multicolumn{1}{l|}{7.31}         
        &                         11.80                      
        &                              10.33              
        & \multicolumn{1}{l|}{13.07}                       
        &                          11.70                    
        \\
        MTRNet~\cite{cao2021latent}
        &  2021 
        &      6.43
        &        9.69
        &       17.08
        & \multicolumn{1}{l|}{22.23}      
        &         8.39
        & \multicolumn{1}{l|}{6.89}         
        &                         11.78                      
        &                              -              
        & \multicolumn{1}{l|}{-}                       
        &                      -                        
        \\
        Refine-Net~\cite{zhou2022refine}
        & 2022
        &      5.92
        &        9.04
        &       16.52
        & \multicolumn{1}{l|}{22.19}      
        &         7.70
        & \multicolumn{1}{l|}{7.20}         
        &                         11.43                      
        &                              18.09              
        & \multicolumn{1}{l|}{19.73}                       
        &                          18.91                    
        \\
        Zhang~\etal~\cite{zhang2022geometry}
        & 2022
        &      5.65
        &        9.19
        &       16.78
        & \multicolumn{1}{l|}{22.93}      
        &         6.68
        & \multicolumn{1}{l|}{6.29}         
        &                         11.25                      
        &                              9.31              
        & \multicolumn{1}{l|}{13.11}                       
        &                          11.21                    
        \\
        AdaFit~\cite{zhu2021adafit}
        & 2021
        &      5.19
        &        9.05
        &       16.45
        & \multicolumn{1}{l|}{21.94}      
        &         6.01
        & \multicolumn{1}{l|}{5.90}         
        &                         10.76                      
        &                              8.39              
        & \multicolumn{1}{l|}{12.85}                       
        &                          10.62                    
        \\
        GraphFit~\cite{li2022graphfit}
        & 2022
        &    4.45  
        &     \textbf{8.74}
        &       \textbf{16.05}
        & \multicolumn{1}{l|}{21.64}      
        &         5.22
        & \multicolumn{1}{l|}{5.48}         
        &                         10.26                      
        &                              -              
        & \multicolumn{1}{l|}{-}                       
        &                      -                        
        \\
        NeAF~\cite{li2023NeAF}
        & 2023
        & 4.20     
        &     9.25   
        &       16.35
        & \multicolumn{1}{l|}{21.74}      
        &         4.89
        & \multicolumn{1}{l|}{4.88}         
        &                         10.22                      
        &                              -              
        & \multicolumn{1}{l|}{-}                       
        &                      -                        
        \\
        HSurf-Net~\cite{li2022hsurf}
        & 2022
        &  4.17    
        &      8.78  
        &       16.25
        & \multicolumn{1}{l|}{21.61}      
        &         4.98
        & \multicolumn{1}{l|}{4.86}         
        &                         10.11                      
        &                              7.55              
        & \multicolumn{1}{l|}{12.23}                       
        &                          9.89                    
        \\
        \midrule
        Ours 
        & -
        & \textbf{3.84}
        &     \textbf{8.74}   
        &       16.10
        & \multicolumn{1}{l|}{\textbf{21.05}}      
        &         \textbf{4.34}
        & \multicolumn{1}{l|}{\textbf{4.51}}         
        &                         \textbf{9.76}                      
        &                             \textbf{6.94}               
        & \multicolumn{1}{l|}{\textbf{11.66}}                       
        &                          \textbf{9.30}                    
        \\
        \bottomrule
        
    \end{tabular}
    \label{tab: pcpnet result}
\end{table*} 
\begin{figure*}[t]
  \centering
  \includegraphics[width=0.85\linewidth]{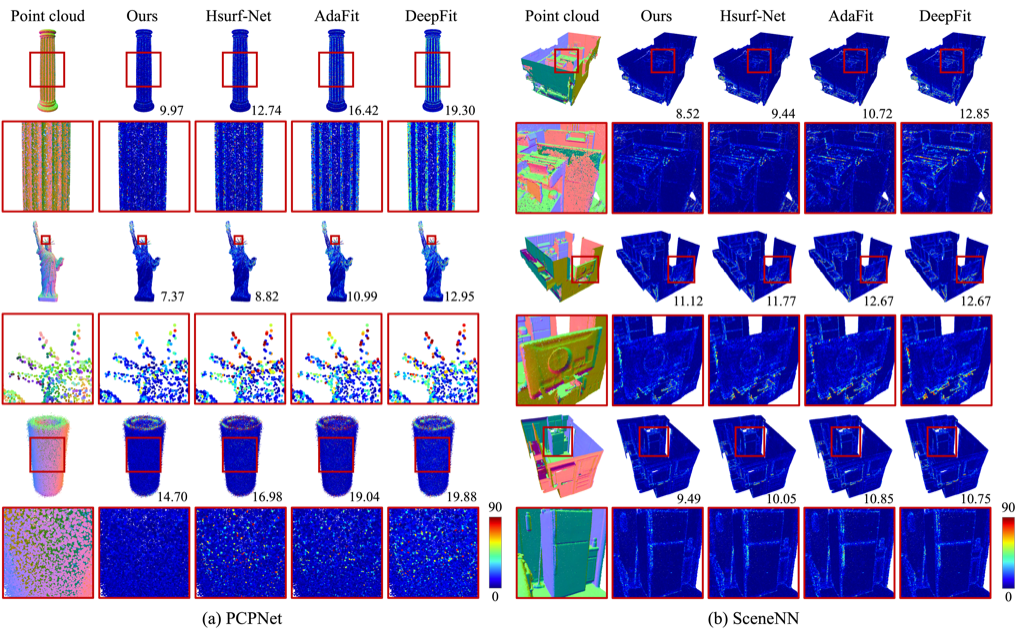}
  \caption{RMSE visualization. Numbers indicate RMSEs. The ``Point cloud'' columns show point clouds colored by ground truth normals. (a) Examples from the PCPNet dataset. From top to bottom: clean, density corrupted, and noisy point clouds. (b) Examples from SceneNN dataset. The top one is clean and the others are noisy point clouds.}
  \Description{Angle error visualizations.}
  \label{fig: error visualizations}
\end{figure*}

\section{Experiment}

\subsection{Datasets and Settings}

We evaluate MSECNet with the PCPNet dataset~\cite{guerrero2018pcpnet} which consists of several synthetic point clouds. 
Each clean cloud has corrupted variants with different noise levels or density variations. Ground truth normals of corrupted variants are the same as clean ones. Some examples are shown in Fig.~\ref{fig: pcpnnet example}. Next, we evaluate the generalizability of MSECNet on the SceneNN dataset~\cite{hua2016scenenn}, a real-world dataset consisting of indoor scenes. We use the dataset prepared by \cite{li2022hsurf}, which contains clean and noisy scenes. All models are pretrained on the PCPNet dataset and applied to the SceneNN dataset without fine-tuning. 
For the PCPNet dataset, we follow the settings in \cite{guerrero2018pcpnet} including the train-test split and data augmentation. During training, we randomly sample 1000 patches for each sample. The patch size is set to $N=700$, following previous works~\cite{zhu2021adafit,li2022hsurf}. The patch size (k) of kNN in the backbone is set to 16 and 9 for the MSEC stream. 
The AdamW~\cite{loshchilov2018decoupled} optimizer with an initial learning rate of $2\times10^{-3}$ is used and the learning rate is decayed by cosine annealing schedule until it is reduced to $2\times10^{-5}$. The model is trained with a batch size of 128 and trained for 150 epochs on four NVIDIA Tesla V100 GPUs.

For the evaluation metric, we use the common angle Root Mean Squared Error (RMSE) between the estimated normal and the ground truth to evaluate performance.

The normal estimation results are reported in Sec.~\ref{sec: normal estimation}. Also, we perform various analyses including the ablation study (Sec.~\ref{sec: ablation}), visualization of edge conditioning (Sec.~\ref{sec: effect of edge conditioning}), efficiency analysis (Sec.~\ref{sec: efficienncy}), and the application to surface reconstruction (Sec.~\ref{sec: surface reconstruction}).

\subsection{Normal Estimation}\label{sec: normal estimation}
As shown in Tab.~\ref{tab: pcpnet result}, MSECNet achieves the best scores in terms of average performance, demonstrating its effectiveness. In addition, MSECNet achieves the best scores in most categories. Thus, MSECNet consistently performs well in a variety of situations.
Compared to baselines, MSECNet performs particularly well when data are significantly corrupted (see ``1.2\%'' and ``Stripes''). We speculate that our fusion strategy and adaptability of the detector lead to a better understanding of the underlying geometry, leading to improved robustness. 
MSECNet also achieves the best score in the real-world scenario, which verifies its good generalizability. 

As shown in Fig.~\ref{fig: error visualizations}~(a), the topmost example shows that explicit edge modeling of MSECNet significantly reduces errors at sharp edges. 
When the density is significantly biased (the middle one), MSECNet produces a robust prediction despite that many points are missing. 
We believe that the fusion of all scales leads to more reliable surface descriptions.
In the bottom example, strong noise causes many false edges to appear, turning a smooth surface into a rough surface. MSECNet produces smooth and accurate predictions, whereas existing methods produce salt-and-pepper errors. This also proves that MSECNet learns the underlying geometry well.

\subsection{Ablation Study}\label{sec: ablation}

\begin{table*}[t]
\small
\caption{Results of ablations. The lower the better. The $\Delta$ column shows the difference of ``Average'' compared with ``Default''.}
\label{tab: ablation}
\begin{tabular}{@{}cl|cccc|cc|c|c@{}}
\toprule
\multicolumn{2}{c|}{\multirow{2}{*}{Ablation}} & \multicolumn{4}{c|}{Noise $\sigma$} & \multicolumn{2}{c|}{Density} & \multicolumn{1}{c|}{\multirow{2}{*}{Average}} 
& \multirow{2}{*}{$\Delta$} 
\\
\cmidrule(lr){3-8}
\multicolumn{2}{c|}{}                          & None    & 0.12\%   & 0.6\%   & 1.2\%   & Stripes      & Gradient      & \multicolumn{1}{c|}{} & \multicolumn{1}{c}{}                         \\ \midrule
Default &
  \begin{tabular}[c]{@{}l@{}}$S$=4, $d_{fused}=1024$, Direct interpolation, \\Space + Channel,  w/ adaptive Lap., w/ edge cond.\end{tabular} & 3.84 & 8.74 & 16.10 & 21.05 & 4.34 & 4.51 & 9.76 & -
   \\
   \midrule
\multirow{4}{*}{(a)} & w/o MSEC stream                  & 5.10  & 8.97  & 16.22 & 21.04  & 5.92  & 5.73 & 10.50 & -0.74\\
                     & $S$=1                  & 3.75 & 9.04 & 16.34 & 21.12 & 4.43 & 4.54 & 9.86 & -0.10\\
                     & $S$=2                  & 3.73 & 8.87 & 16.25 & 21.14 & 4.41 & 4.42 & 9.80 & -0.04\\
                     & $S$=3                  & 3.80 & 8.83 & 16.24 & 21.06 & 4.44 & 4.44 & 9.80 & -0.04 \\
                     \midrule
\multirow{2}{*}{(b)} & $d_{fused}$=256      & 3.97 & 8.81 & 16.10 & 20.98 & 4.64 & 4.67 & 9.86 & -0.10\\
                     & $d_{fused}$=512      & 3.85 & 8.75 & 16.07 & 21.04 & 4.46 & 4.58 & 9.79 & -0.03\\
                     \midrule

\multirow{1}{*}{(c)} 
& Gradual interpolation
& 3.91
& 8.92
& 16.15
& 21.12 
& 4.59
& 4.58
& 9.88 
& -0.12 \\

                     \midrule
\multirow{2}{*}{(d)} & Channel transformation         & 4.03 & 8.82 & 16.12 & 21.09 & 4.71 & 4.75 & 9.92 & -0.16 \\
& Space transformation           & 3.89 & 8.83 & 16.16 & 21.04 & 4.52 & 4.55 & 9.83 & -0.07 \\

                     \midrule
\multirow{1}{*}{(e)} & w/o adaptability & 3.86 & 8.85 & 16.17 & 21.05 & 4.44 & 4.52 & 9.82 & -0.06 \\
                     \midrule
\multirow{2}{*}{(f)} 
& edge only & 4.37 & 8.89 & 16.22 & 21.09 & 5.10 & 5.01 & 10.11 & -0.35\\ 
& w/o conditioning & 3.87 & 8.84 & 16.11 & 21.07 & 4.48 & 4.53 & 9.82 & -0.06\\
                     
                      
   \bottomrule
                     
\end{tabular}
\end{table*}

\indent 

\textbf{(a) Number of scales.}
The effectiveness of the MSEC stream is investigated by varying the number of scales involved, where $S=1$ represents the smallest scale and each increment of $S$ adds an adjacent larger one. As shown in Tab.~\ref{tab: ablation}~(a), the model achieves mediocre performance when the MSEC stream is not used. However, adding just one scale significantly improves performance, demonstrating the effectiveness of explicit edge modeling.
Also, consistent improvement for noisy samples as $S$ increases from 1 to 4 suggests that integrating more scales provides better robustness to noise.

\textbf{(b) Size of $d_{fused}$.} 
The amount of information retained from different scales is determined by the size of $d_{fused}$. Tab.~\ref{tab: ablation}~(b) shows that increasing $d_{fused}$ from 256 to 1024 results in a reduction in the average error, indicating that a larger $d_{fused}$ can preserve more useful information. The performance of ``None'' and density variations also improves as $d_{fused}$ increases. Therefore, a larger $d_{fused}$ can provide a more comprehensive descriptor and enhance the network's robustness against sampling biases.

\textbf{(c) Interpolation strategy.}
We investigate the effectiveness of our direct interpolation strategy by comparing it to the standard gradual interpolation, i.e., the feature propagation layer~\cite{qi2017pointnet++}. Note that no transformation is applied for a fair comparison. As shown in Tab.~\ref{tab: ablation}~(c), our direct interpolation results in lower errors consistently across all corruption types.

\textbf{(d) Space and channel transformations.} 
As shown in Tab.~\ref{tab: ablation}~(d), an increase in average error indicates that a channel transformation cannot make full use of multi-scale features. The significant decrease in estimation under density variations indicates that sampling biases cannot be well resolved. As expected, the situation improves when the space transformation is used instead of the channel one. The combination of space and channel transformations achieves the best performance, which fully verifies their necessities.  

\textbf{(e) Effect of the adaptability.} 
To validate the effect of the adaptability in the adaptive Laplacian, we remove $\theta$ and $\varphi$ from Eq.~(\ref{eq: adaptive laplacian}). As shown in Tab.~\ref{tab: ablation}~(e), the adaptability is effective for estimation under low to medium levels of noise and ``Striped'' corruption. Thus, the adaptability is most effective in situations where surface geometry is significantly distorted. Also, no improvement is seen under the strongest noise level. For complex point clouds, it is assumed that the noise almost completely destroys the raw local geometry, making edge detection challenging. 

\textbf{(f) Effect of the edge conditioning.}
As shown in Tab.~\ref{tab: ablation}~(f), the importance of global information
is confirmed by the 0.35 drop in performance when $\textbf{f}_i^{b}$ is removed from Eq.~(\ref{eq: edge conditioning}). To verify the effect of edge conditioning, we only remove $\textbf{f}_i^{b}$ from $\gamma$, which leads to a performance drop of 0.06. This result shows that conditioning is useful, but the degradation is not as severe as when $\textbf{f}_i^{b}$ is entirely removed thanks to the residual connection.

\subsection{Visualization of Edge Conditioning}\label{sec: effect of edge conditioning}
\begin{figure}[t]
    \centering
    \includegraphics[width=0.7\linewidth]{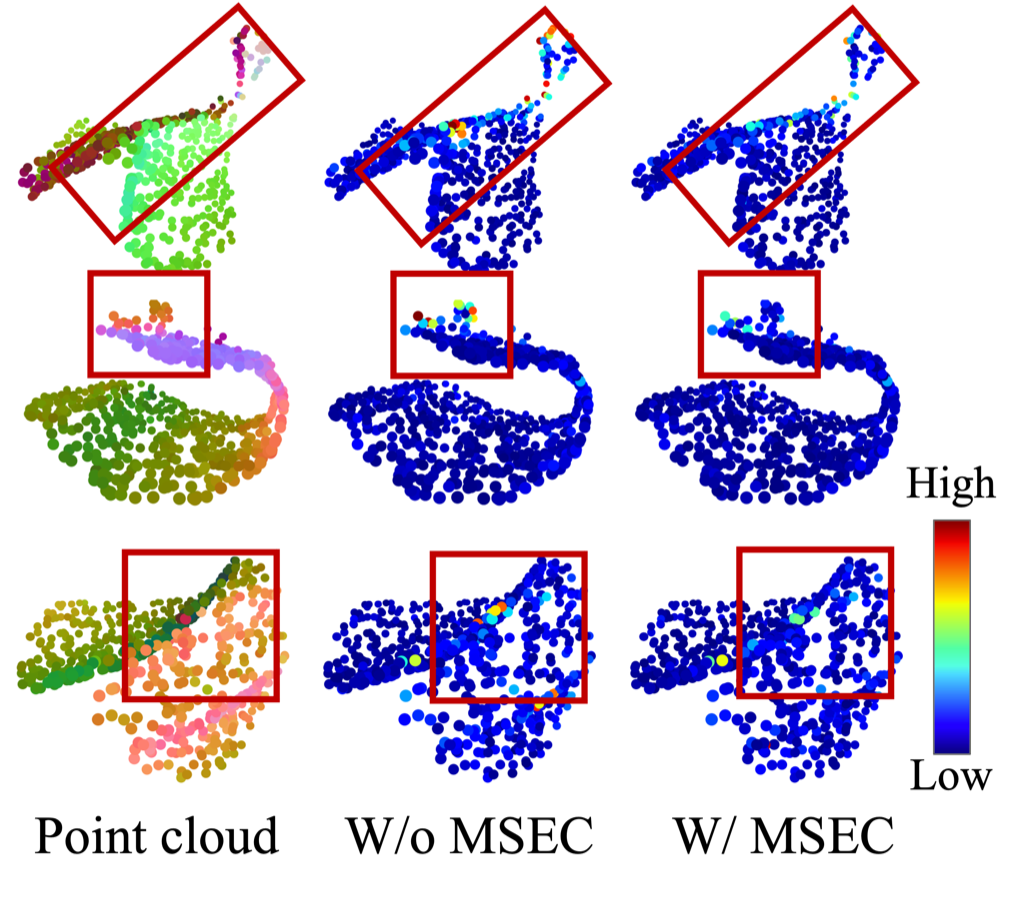}
    \caption{Effect of the MESC stream on RMSEs.}
    \label{fig: with without error}
\end{figure}
\begin{figure}[t]
    \centering
    \includegraphics[width=0.8\linewidth]{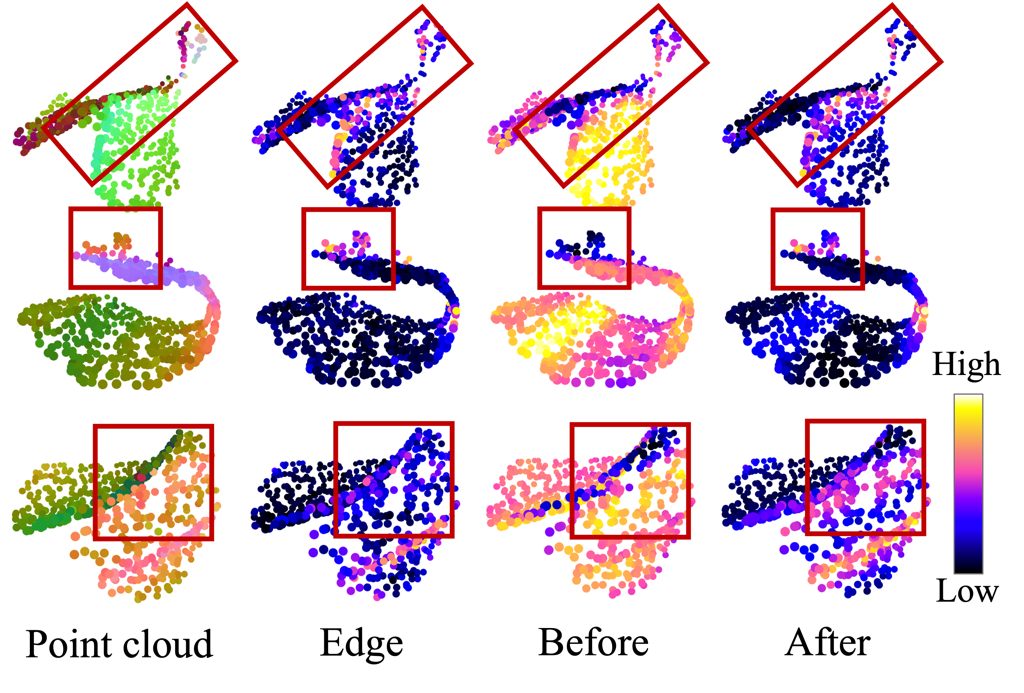}
    \caption{Visualization of learned features. The sample patches are the same ones as Fig.~\ref{fig: with without error}. ``Edge'' indicates detected edges. ``Before'' shows the output of the backbone ($\textbf{f}_i^b$) and ``After'' shows the features after edge conditioning.}
    \label{fig: with without feature}
\end{figure}
\begin{table*}[tb]
\small
\caption{Number of parameters and inference Time. The average inference time for point clouds with 100k points is reported.  }
\begin{tabular}{@{}lllllll@{}}
\toprule
              & Ours ($d_{fused}=256$) & Ours ($d_{fused}=512$) & Ours ($d_{fused}=1024$) & HSurf-Net~\cite{li2022hsurf} & AdaFit~\cite{zhu2021adafit} & DeepFit~\cite{ben2020deepfit} \\ \midrule
Param. (M) & \textbf{1.92} & 3.83 & 10.40 & 2.16      & 4.87   & 3.53    \\
Time (s)   & \textbf{0.45} & \textbf{0.45} & 0.93 & 56.70     & 26.54         & 45.12         \\ \bottomrule
\end{tabular}
\label{tab: efficiency}
\end{table*}
As shown in Fig.~\ref{fig: with without error}, high RMSEs are observed mainly at the edges, where errors are corrected by the MSEC stream. Thus, the effect of the MSEC stream is most pronounced in the accurate estimation at the edges.
Further, using the same sample as in Fig.~\ref{fig: with without error}, we visualize the detected edges and the effect of edge conditioning using $l^2$-norm. As can be seen from Fig.~\ref{fig: with without feature}, the edges are correctly detected, as strong responses are obtained in the edge regions. After conditioning, the feature distribution changes significantly, confirming that edge-conditioned representation is obtained. Combining Fig.~\ref{fig: with without error} and \ref{fig: with without feature}, we confirm that the MSEC stream contributes to the improvement.

\subsection{Number of Parameters and Inference Time}\label{sec: efficienncy}
The results are presented in Tab.~\ref{tab: efficiency}. We use an NVIDIA Tesla V100 with a batch size of 128. As shown, our default model with $d_{fused}=1024$ is significantly faster than other methods. Additionally, we can significantly reduce the model size by decreasing $d_{fused}$ without compromising performance too much (see Tab.~\ref{tab: ablation}~(b)). Therefore, we conclude that MSECNet can be a practical choice for normal estimation in situations where resources are limited.

\subsection{Application to Surface Reconstruction}\label{sec: surface reconstruction}

\begin{figure}[tb]
  \centering
  \includegraphics[width=0.9\linewidth]{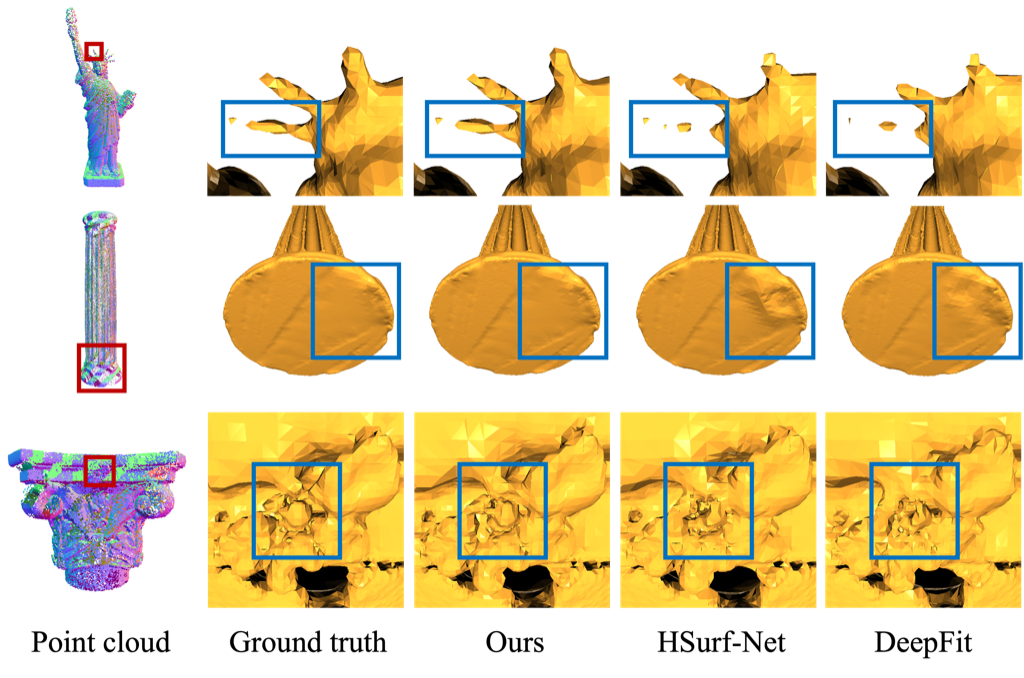}
  \caption{Result of surface reconstruction.}
  \Description{}
  \label{fig: reconstruction}
\end{figure}


We use surface reconstruction to demonstrate the utility of MSECNet. Poisson reconstruction~\cite{kazhdan2006poisson} is used to reconstruct surfaces. Fig.~\ref{fig: reconstruction} shows that the reconstructed surfaces using our normals are more precise than those produced by other methods.

\section{Conclusion}
We propose MSECNet, a novel method for improving normal estimation at normal varying regions. The key is that the modeling of normal variations is converted to the modeling of geometric edges. We design the MSEC stream that performs multi-scale edge conditioning. MSECNet achieves state-of-the-art performance on both synthetic and real-world datasets while running significantly faster than existing methods. We also confirm that the MSEC stream better models the geometry in the presence of data corruptions. Therefore, a promising direction for future research is to extend MSECNet to tasks such as point cloud filtering and generation.  

\begin{acks}
This paper is based on the results of a project, JPNP20006, commissioned by the New Energy and Industrial Technology Development Organization (NEDO). We would also like to acknowledge the support from JSPS Grant-in-Aid for Scientific Research (21K12042).
\end{acks}

\clearpage
\bibliographystyle{ACM-Reference-Format}
\balance
\bibliography{sample-base}










\end{document}


\title{Supplementary Materials for \\MSECNet: Accurate and Robust Normal Estimation for 3D Point Clouds by Multi-Scale Edge Conditioning}

\renewcommand{\shortauthors}{Haoyi Xiu, Xin Liu, Weimin Wang, Kyoung-Sook Kim, \& Masashi Matsuoka}
\newcommand{\etal}{\textit{et al.}}
\newcommand{\R}{\mathbb{R}}

\begin{abstract}
We provide additional details and analysis in this material. Moreover, we have included a video that contains animated versions of the figures in the main paper.   
\end{abstract}

\ccsdesc[500]{Computing methodologies~Shape representations}
\ccsdesc[500]{Computing methodologies~Scene understanding}

\copyrightyear{2023}
\acmYear{2023}
\setcopyright{acmlicensed}\acmConference[MM '23]{Proceedings of the 31st
ACM International Conference on Multimedia}{October 29-November 3,
2023}{Ottawa, ON, Canada}
\acmBooktitle{Proceedings of the 31st ACM International Conference on
Multimedia (MM '23), October 29-November 3, 2023, Ottawa, ON, Canada}
\acmPrice{15.00}
\acmDOI{10.1145/3581783.3613762}
\acmISBN{979-8-4007-0108-5/23/10}

\begin{teaserfigure}
  \includegraphics[width=0.9\textwidth]{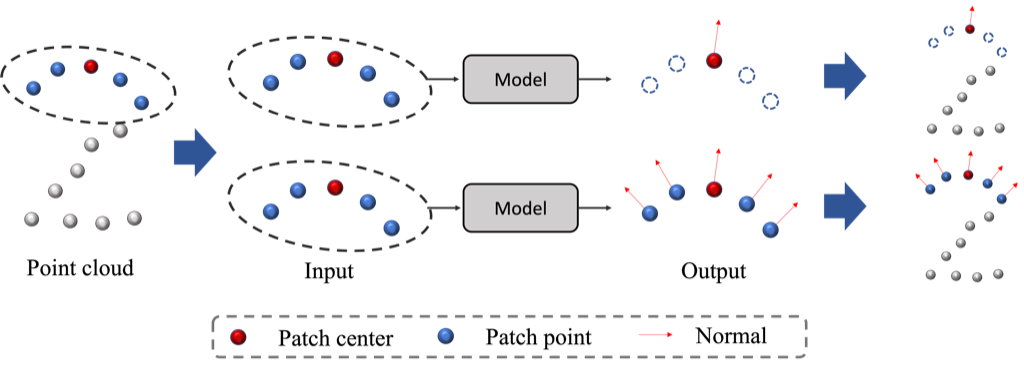}
  \caption{Comparison of different architectures. Top: the architecture that produces a normal vector for the patch center. Bottom: the architecture that produces normal vectors for all input points.}
  \Description{.}
  \label{fig: inference comparison}
\end{teaserfigure}


\maketitle


\section{Additional Implementation Details and Settings}
\paragraph{Input transformation} We perform Principal Component Analysis (PCA) on the input patch before it is fed into the model. This transformation aligns the axes of the patch with its principal components. The purpose of this step is to potentially enhance performance by simplifying training through the reduction of input patch variations. This technique is also adopted in previous works (e.g., \cite{li2022hsurf}). 
\paragraph{Dimensions of transformations} $\phi$ defined in Eq.~(1) merges features of differnet scales into a vector. Four scales are involved, with dimensions of 64, 128, 256, and 512, respectively. As we set $d_{fused}$ as 1024, $\phi$ maps a 960-dimensional vector to 1024 one. $\alpha$ is a mapping from 1027 to 1024 dimensions, and $\beta, \varphi,\theta$ are mappings from 1024 to 1024 dimensions. As for $gamma$, as we set $d_{out}=128$, it is a mapping from 1152 to 128 dimensions. The estimated normal ($\widehat{\textbf{n}}_i$) is obtained with an MLP which maps a 128-dimensional vector to a 3-dimensional one. 
\paragraph{How the features are visualized in Fig.~6 We convert the vector feature into a scalar one by $l^2$-norm. Suppose that $\textbf{w}_i=(w_{i,1}, w_{i,2}, ..., w_{i,d})$ denotes a $d$ dimensional feature of point $i$. The scalar feature is computed as $v_i = \sqrt{(w_{i,1})^2+ (w_{i,2})^2 ... +(w_{i, d})^2}$ where $v_i$ denotes the scalar feature used for visualization. We adopt $l^2$-norm because its computation involves all individual components, thus showing the overall characteristics.
\paragraph{Settings of surface reconstruction} We adopt Poisson reconstruction~\cite{kazhdan2006poisson} algorithm for surface reconstruction. Its implementation is provided by Open3D~\cite{zhou2018open3d}. In our case, the depth of the octree is set to 9. 

\paragraph{Details of architecture design and inference} 
Fig~\ref{fig: inference comparison} illustrates a comparison between the commonly used architecture type and our architecture. The top architecture, as seen in previous works (e.g., \cite{guerrero2018pcpnet,ben2020deepfit,zhu2021adafit,li2022hsurf}), takes a patch as input and generates a single normal vector for the patch center. This approach requires visiting all points in the input point cloud to infer normal vectors. In contrast, we employ the bottom architecture where normals for all patches are estimated. By ensuring that the sampled patches cover the entire point cloud, we can infer normals for all points efficiently. To achieve this, we use potential-based sampling to generate patch sets that cover the point cloud. This sampling method updates the potential of each point and its neighbors during the sampling process, ensuring that the next patch center is always a different point. However, the potential of points near the patch boundaries remains unchanged since they may produce unreliable predictions due to a lack of contextual information. As a result of our approach, every point in the point cloud is sampled at least once, while some points may be sampled multiple times. When multiple predictions exist for a point, we average them to obtain a final prediction. This method enhances the efficiency of inference while ensuring comprehensive coverage of the point cloud.
\section{Additional Analysis}
\subsection{The Number of Scales and Fusion Dimension}
\begin{table}[t]
    \caption{Effect of $S$ and $d_{fused}$}
    \centering
    \begin{tabular}{l|ccc|c}
         \toprule
         \diagbox[width=7em]{$S$}{$d_{fused}$}
         & 256
         & 512
         & 1024
         & Row average
         \\
         \midrule
         1
         & 9.87
         & 9.83
         & 9.86
         & 9.85
         \\
         2
         & 9.81
         & 9.82
         & 9.80
         & 9.84
         \\
         3
         & 9.87
         & 9.79
         & 9.80
         & 9.82
         \\
         4
         & 9.86
         & 9.79
         & 9.76
         & 9.80
         \\
         \bottomrule
         
    \end{tabular}
    
    \label{tab: s and d_fusion}
\end{table}
The additional analysis concerning the number of scales ($S$) and the fusion dimension ($d_{fused}$) is presented. The result is listed in \cref{tab: s and d_fusion}. The progressive improvement shown in the ``Row average'' column indicates that the increase in the number of scales improves overall performance. 
On the other hand, the peak of performance appears at different $S$ for different $d_{fused}$. More specifically, the peak of $d_{fused}=256$ is reached when $S=2$, while for $d_{fused}=512$ and $d_{fused}=1024$ are $S=3 / 4$ and $S=4$, respectively. This shows that a sufficient size of $d_{fused}$ is necessary to better utilize information from different scales. We conjecture that information loss occurs if $d_{fused}$ is not big enough to contain all information from different scales.

\subsection{Comparison of the Adaptive Laplacian and EdgeConv}
\begin{table}[t]
    \centering
    \caption{Comparison of the Adaptive Laplacian and EdgeConv~\cite{wang2019dynamic}. $\textbf{x}_i$ represents the feature of point $i$. $\mathcal{N}(i)$ and $\mathcal{M}(i)$ denote the 3D and feature space neighborhood of point $i$, respectively. Both $\alpha$ and $\beta$ are MLPs. ``Average'' denotes the angle RMSE.}
    \begin{tabular}{l|l}
        \toprule
         Method
         & Equation 
         \\
         \midrule
         Ada. Lap.
         & $\textbf{x}_i' = \beta\left(\frac{1}{|\mathcal{N}(i)|}\sum_{j \in \mathcal{N}(i)}\alpha\left(\textbf{x}_j - \textbf{x}_i\right)\right)$
         \\
         EdgeConv
         &  $\textbf{x}_i' = \max_{j\in\mathcal{M}(i)} \alpha\left(\textbf{x}_i\right) + \beta\left({\textbf{x}_j - \textbf{x}_i}\right)$
         \\
         \bottomrule
    \end{tabular}
    \label{tab: sup_edgeconv_adaptive_laplcian}
\end{table}
We compare the proposed Adaptive Laplacian with EdgeConv, the core operator of DGCNN~\cite{wang2019dynamic}, to better clarify our design choices. The expressions of them are shown in Tab.~\ref{tab: sup_edgeconv_adaptive_laplcian}. While they look somewhat similar to each other, we believe that they differ significantly in their purposes. We design the Adaptive Laplacian for edge detection. Since the Laplacian is a widely used method for edge detection, we believe it is reasonable to design an edge detection operator based on it. On the other hand, EdgeConv is designed for feature aggregation, whose purpose is to learn a more discriminative representation by aggregating features from different neighborhoods.

The difference in purpose leads to the difference in mathematical designs. The first difference is the use of the global feature $\textbf{x}_i$. EdgeConv must include the global feature because it is designed for feature aggregation. Without global information, the operator cannot learn useful features for the task at hand (e.g., classification). On the other hand, Adaptive Laplacian is designed to detect edges, which are local features. Therefore, the inclusion of global features needs to be avoided. The second difference is the definition of the neighborhood. EdgeConv searches for neighbors in the feature space to enlarge the receptive fields. Therefore, it captures the long-range relationship instead of local properties, which makes the resulting features more suitable for high-level tasks. In contrast, the adaptive Laplacian performs kNN in 3D space because edges are spatially localized.

\subsection{The effect of the MSEC stream on DGCNN}

\begin{table}[t]
    \centering
    \caption{The effect of the MSEC stream on DGCNN~\cite{wang2019dynamic}. ``Average'' denotes the average angle RMSE.}
    \begin{tabular}{l|c}
        \toprule
         Method & Average \\
         \midrule
         DGCNN~\cite{wang2019dynamic}
         & 12.43\\
         DGCNN + MSEC stream
         & 10.40\\
         \bottomrule
    \end{tabular}
    \label{tab: sup_dgcnn}
\end{table}
In this section, we examine the effect of the MSEC stream on another backbone. By default, we use PointNet++~\cite{qi2017pointnet++} because it efficiently encodes the multi-scale features through subsampling. To study its effect on the network without subsampling, we integrate the MSEC stream into DGCNN~\cite{wang2019dynamic}. The core operator of DGCNN is EdgeConv. EdgeConv potentially captures different multi-scale structures of point clouds compared to PointNet++ by searching for neighbors in the feature space. We treat the output of each EdgeConv as a feature of a particular scale and feed it into the multi-scale fusion module of the MSEC stream. Note that no interpolation is performed because the resolution of the points is kept the same in DGCNN. The results are shown in Tab.~\ref{tab: sup_dgcnn}. We observe a significant improvement by integrating the MSEC stream into DGCNN. Therefore, the MSEC stream is effective even if its backbone does not perform subsampling to explicitly encode multi-scale structures.


\section{Additional Visualization}
\subsection{Video}
We provide a video that contains animated versions of Fig.~4, Fig.~5, Fig.~6, and Fig.~7 to better demonstrate the details of the 3D point clouds. 

\subsection{Results on Semantic3D}
In addition, we conduct experiments to verify the robustness of our method to sensor variations and data type. Specifically, we used Semantic3D~\cite{hackel2017semantic3d}, an outdoor dataset acquired by a terrestrial LiDAR sensor. Due to the lack of ground truth normals, we only visualize the estimation results. As shown in Fig.~\ref{fig: sup_semantic3d}, we find that our method can produce a more accurate estimation in normal varying regions compared to other methods, which is consistent with the results presented in the paper. Therefore, our method is robust to different sensors and data types.
\begin{figure}[t]
    \centering
    \includegraphics[width=\linewidth]{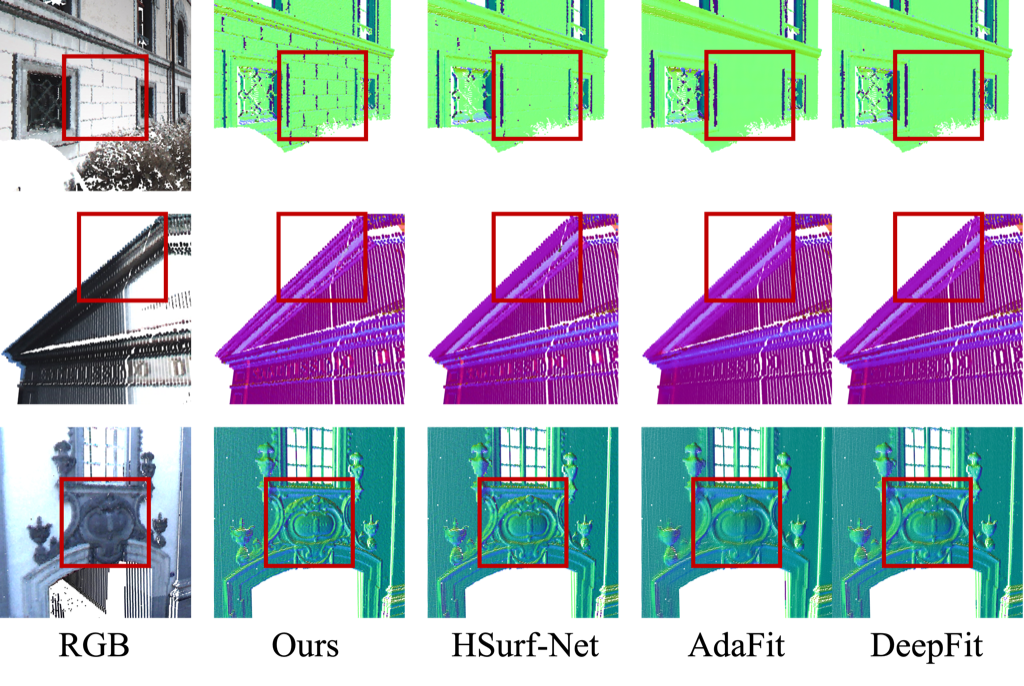}
    \caption{Results of normal estimation on the Semantic3D dataset.}
    \label{fig: sup_semantic3d}
\end{figure}

\clearpage
\bibliographystyle{ACM-Reference-Format}
\balance
\bibliography{sample-base}